\documentclass[10pt, a4paper, twocolumn, teaser, showabstract]{naverlabseurope}

\usepackage{lipsum}
\usepackage{multicol}
\usepackage{tikz,tkz-kiviat,pgfplots}

\usepackage{makecell}
\usepackage{multirow}
\usepackage{svg}
\usepackage{caption}

\usepackage{graphicx}
\usepackage{floatrow}
\usepackage{tcolorbox}
\usepackage{xcolor}
\usepackage{url}
\usepackage{adjustbox}
\usepackage{tabularx}
\usepackage{booktabs}
\usepackage[justification=centering]{subfig}

\graphicspath{{figures/}}

\title{Speech-MASSIVE: A Multilingual Speech Dataset for SLU and Beyond}

\correspondingauthor{beomseok.lee@unitn.it}

\authors{Beomseok Lee¹²³, Ioan Calapodescu², Marco Gaido³, Matteo Negri³, Laurent Besacier²}
\affiliations{¹University of Trento, Italy\\²NAVER LABS Europe, France\\³Fondazione Bruno Kessler, Italy}

\begin{abstract}
We present Speech-MASSIVE, a multilingual Spoken Language Understanding (SLU) dataset comprising the speech counterpart for a portion of the MASSIVE textual corpus. 
Speech-MASSIVE covers 12 languages from different families and inherits from MASSIVE the annotations for the intent prediction and slot-filling tasks.
Our extension is prompted by the scarcity of massively multilingual SLU datasets and the growing need for versatile speech datasets to assess foundation models (LLMs, speech encoders) across languages and tasks. 
We provide a multimodal, multitask, multilingual dataset and report SLU baselines using both cascaded and end-to-end architectures in various training scenarios (zero-shot, few-shot, and full fine-tune).
Furthermore, we demonstrate the suitability of Speech-MASSIVE for benchmarking other tasks such as speech transcription, language identification, and speech translation.
The dataset, models, and code are publicly available at: \url{https://github.com/hlt-mt/Speech-MASSIVE}
\end{abstract}

\begin{document}

\maketitle

\section{Introduction}
\label{introduction}

Multilingual speech corpora have limited coverage of speech-related tasks, primarily focusing on automatic speech recognition (ASR) \cite{commonvoice,fleurs,MaSS,multilinguallibrispeech} and speech translation (ST) \cite{mustc,mtedx,europarlst,covost2}, while neglecting spoken language understanding (SLU -- the task of extracting semantic information from spoken utterances, which typically involves subtasks like intent detection and slot filling).
Unlike text processing, where extensive efforts in natural language understanding (NLU) have led to resources covering a wide range of languages \cite{mlqa,multi3nlu,multiatis,massive}, SLU datasets are mainly English-centric \cite{slurp}, with few exceptions \cite{snips,portmedia,italic}.

Our goal is to bridge the gap in  multilingual SLU drawing inspiration from \cite{italic} and collecting speech recordings in multiple languages. We start with the  MASSIVE NLU (i.e. textual) dataset \cite{massive}, an ideal foundation due to its size, domain diversity, and broad coverage of languages, intent, and slot types. Developed by commissioning professional translators to localize the English SLURP dataset \cite{slurp} into 51 languages, MASSIVE comprises 1M labeled utterances spanning 18 domains, with 60 intents and 55 slots. Our contribution, Speech-MASSIVE, spans 12 languages from diverse families: Arabic, German, Spanish, French, Hungarian, Korean, Dutch, Polish, European Portuguese, Russian, Turkish, and Vietnamese. It also facilitates evaluation across various speech tasks beyond SLU, including ASR, ST, and language identification (LID).
We release Speech-MASSIVE publicly under CC-BY-NC-SA 4.0 license.\footnote{\url{https://hf.co/datasets/FBK-MT/Speech-MASSIVE}}

Besides detailing the creation process involving a crowdsourcing-based protocol for data collection and quality control, this paper presents
baseline SLU results on Speech-MASSIVE. Our results with both cascade and end-to-end architectures trained in different conditions (zero-shot, few-shot,  full fine-tune) will enable future comparisons and tracking SLU advancements compared to the more mature field of NLU. Lastly, we showcase Speech-MASSIVE's versatility through additional experiments on ASR, LID, and ST.

\section{Speech-MASSIVE}
\label{speech-massive}

\subsection{Data collection and validation process}
\label{subsec:summary_data_collection}

We created the speech counterpart of textual MASSIVE data by recruiting native speakers through the Prolific crowdsourcing platform.\footnote{\url{https://www.prolific.com}, Compensated £9 per hour.} 
A first group of workers was instructed to record the spoken version of MASSIVE sentences with guidelines emphasizing the importance of accurate and natural reading, as well as proper recording conditions and strict adherence to the corresponding text.
To ensure high final data quality, a second group of native speakers validated the recorded utterances. During validation, participants were directed to read the original text, listen to the recording, and label it as \textit{valid} or \textit{invalid}.
Those marked as invalid underwent a second iteration of this two-step (recording and validation) process. 
After the second iteration, the process concluded, irrespective of the outcome of the second validation phase, to avoid potentially endless cycles. 
This decision was also informed by the observation that, upon inspecting the invalid recordings, we found some were marked as such not due to a lack of adherence of the speech to the text but because of grammatical errors in the original MASSIVE dataset text. Correcting these errors was beyond the scope of our work.

To further enhance the reliability of the collected dataset, we implemented two additional precautions. 
During the recording phase, we instructed participants to review their own recordings before proceeding to the next sample, allowing them to re-record if the audio was not properly acquired.
Additionally, in the validation step, four speech utterances were chosen from Common Voice \cite{commonvoice} and inserted among the samples for validation. 
Out of these four quality control samples, two intentionally featured audio-transcript mismatches to be marked as invalid. 
The other two cases had perfect  audio-transcript alignment to be marked as valid. 
Care was taken to select quality control samples with clear and intelligible audio. 
Validation results from  a  Prolific user were retained only if they accurately assessed all four quality control samples. 
Any mistakes led to the disregarding of their validations, requiring the entire set of samples from that user to be re-validated by other participants.

\begin{table}
    \resizebox{\linewidth}{!}{
\begin{tabular}{c|c|c|c|c|c|c|c}\toprule
lang 
&split &\makecell{\# sample} &\makecell{\# valid} &\makecell{\# hrs} &\makecell{total \\\# spk (M/F/U)} &WER &CER\\\hline
\multirow{2}{*}{ar} 
&train-115 &115 &115 &0.14 &8 (4/4/0) &- &- \\
&dev &2033 &2027 &2.12 &36 (22/14/0) &31.75 &14.43 \\
&test &2974 &2962 &3.23 &37 (15/17/5) &34.19 &15.85 \\
\hline
\multirow{3}{*}{de} 
&train-115 &115 &115 &0.15 &7 (3/4/0) &- &- \\
&train-full &11514 &11201 &12.61 &117 (50/63/4) &- &- \\
&dev &2033 &2032 &2.33 &68 (35/32/1) &11.24 & 3.96 \\
&test &2974 &2969 &3.41 &82 (36/36/10) &11.84 & 4.16 \\
\hline
\multirow{2}{*}{es}
&train-115 &115 &115 &0.13 &7 (3/4/0) &- &- \\
&dev &2033 &2024 &2.53 &109 (51/53/5) &7.61 &3.00 \\
&test &2974 &2948 &3.61 &85 (37/33/15) &8.95 &3.76 \\
\hline
\multirow{3}{*}{fr}
&train-115 &115 &115 &0.12 &103 (50/52/1) &- &- \\
&train-full &11514 &11481 &12.42 &103 (50/52/1) &- &- \\
&dev &2033 &2031 &2.20 &55 (26/26/3) &10.20 &4.42 \\
&test &2974 &2972 &2.65 &75 (31/35/9) &11.09 &4.71 \\
\hline
\multirow{2}{*}{hu}
&train-115 &115 &115 &0.12 &8 (3/4/1) &- &- \\
&dev &2033 &2019 &2.27 &69 (33/33/3) &25.96 &10.93 \\
&test &2974 &2932 &3.30 &55 (25/24/6) &20.98 &6.01 \\
\hline
\multirow{2}{*}{ko}
&train-115 &115 &115 &0.14 &8 (4/4/0) &- &- \\
&dev &2033 &2032 &2.12 &21 (8/13/0) &25.29 &7.13 \\
&test &2974 &2970 &2.66 &31 (10/18/3) &26.42 &8.04 \\
\hline
\multirow{2}{*}{nl}
&train-115 &115 &115 &0.12 &7 (3/4/0) &- &- \\
&dev &2033 &2032 &2.14 &37 (17/19/1) &11.03 &3.98 \\
&test &2974 &2959 &3.30 &100 (48/49/3) &10.52 &3.82 \\
\hline
\multirow{2}{*}{pl}
&train-115 &115 &115 &0.10 &7 (3/4/0) &- &- \\
&dev &2033 &2024 &2.24 &105 (50/52/3) &9.94 &4.88 \\
&test &2974 &2933 &3.21 &151 (73/71/7) &12.58 &6.22 \\
\hline
\multirow{2}{*}{pt}
&train-115 &115 &115 &0.12 &8 (4/4/0) &- &- \\
&dev &2033 &2031 &2.20 &107 (51/53/3) &11.73 &5.10 \\
&test &2974 &2967 &3.25 &102 (48/50/4) &12.11 &5.13 \\
\hline
\multirow{2}{*}{ru}
&train-115 &115 &115 &0.12 &7 (3/4/0) &- &- \\
&dev &2033 &2032 &2.25 &40 (7/31/2) &8.55 &4.06 \\
&test &2974 &2969 &3.44 &51 (25/23/3) &8.99 &4.57 \\
\hline
\multirow{2}{*}{tr}
&train-115 &115 &115 &0.11 &6 (3/3/0) &- &- \\
&dev &2033 &2030 &2.17 &71 (36/34/1) &16.65 &4.56 \\
&test &2974 &2950 &3.00 &42 (17/18/7) &18.06 &5.05 \\
\hline
\multirow{2}{*}{vi}
&train-115 &115 &115 &0.11 &7 (2/4/1) &- &- \\
&dev &2033 &1978 &2.10 &28 (13/14/1) &16.65 &10.5 \\
&test &2974 &2954 &3.23 &30 (11/14/5) &14.94 &9.77 \\
\bottomrule
    \end{tabular}
}
    \caption{
    Speech-MASSIVE's overall statistics. `\# hrs' displays the recording duration for all samples (including invalid), while `\# spk (Male/Female/Unknown)' indicates the number of speakers for all the samples (including invalid). The last 2 columns (`WER', and `CER') measures Whisper ASR performance.
    }
    \label{tab:datatable}
\end{table}

\subsection{Overall statistics}

We chose 12 languages based on various criteria. Initially, we considered the number of registered users on  Prolific, sorting the 51 languages covered in MASSIVE.
Languages with fewer than 200 users were excluded to ensure sufficient worker participation to complete the entire acquisition and validation process in reasonable time.
Italian was also excluded due to the availability of the full dataset elsewhere \cite{italic}.
Finally, with an eye at the balance between budget considerations and linguistic diversity, from the remaining 18 languages we selected Arabic, German, Spanish, French, Hungarian, Korean, Dutch, Polish, European Portuguese, Russian, Turkish, and Vietnamese.

We collected speech recordings for MASSIVE's development and test splits. Acquiring the full training dataset (11,514 utterances for each of the 12 languages) exceeded our budget.
In a concession, our emphasis was placed on acquiring comprehensive training data for French and German, while we obtained limited few-shot training data consisting of 115 utterances from the training set for the remaining 10 languages (\textit{train-115} split).

Columns 1-6 of Table \ref{tab:datatable} provide statistics for the collected dataset, including, for each language, the available data splits, the number of recordings, hours of speech, and speakers (total, male, female and unknown).
The ``\# valid'' column indicates the count of human-validated utterances for each data split after the two iterations. 
As a few speech recordings remained invalidated after our two recording-validation cycles, we retained for each utterance the candidate with the lowest Word Error Rate (WER) as transcribed using Whisper \cite{whisper}. 
This ensures speech availability for all MASSIVE utterances, even if some may not perfectly align with the reference transcript.
Additional information regarding this is included in the corpus metadata.
\begin{figure*}[t]
     \centering
     \makebox[\textwidth][c]{\includegraphics[width=1.0\textwidth]{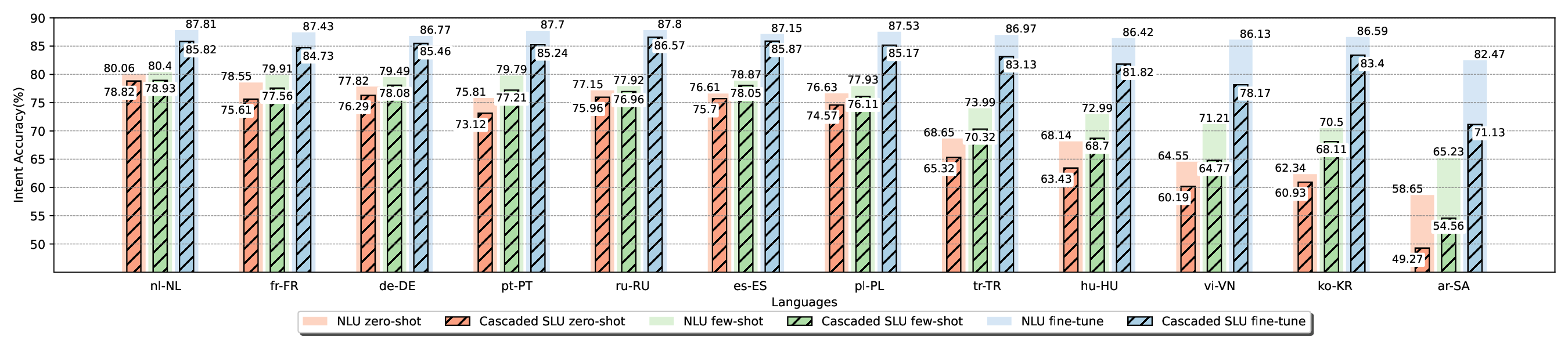}}
     \caption{NLU vs Cascaded SLU (Intent Accuracy) on our Speech-MASSIVE Dataset.}
     \label{fig:NLU-SLU}
\end{figure*}
\subsection{ASR assessment}
\label{ASR_model_intro}
To assess Speech-MASSIVE in multilingual ASR, we used Whisper, since it is one of the recent state-of-the-art multilingual speech recognition models.
We selected Whisper-large-v3,\footnote{\url{https://hf.co/openai/whisper-large-v3}} utilizing it without additional fine-tuning for our ASR evaluation. Table \ref{tab:datatable} shows WER and character error rate (CER) across languages and data splits.
We compared ASR error rates to those obtained on the FLEURS dataset \cite{fleurs}.\footnote{Accessible for our 12 languages except Arabic at \url{https://github.com/openai/whisper/discussions/1762}}  FLEURS generally yields lower WERs/CERs compared to Speech-MASSIVE. The same observation was made for Italian in \cite{italic}, which followed a recording methodology similar to ours. 
This suggests that the higher WERs are likely due to the inherent difficulty of MASSIVE utterances compared to those in FLEURS. Furthermore, there are still discrepancies between our Whisper model's hypotheses and the references in the MASSIVE dataset (e.g., numbers reported in letters in MASSIVE references), which we did not address as optimizing ASR WER was not our main goal.
Finally, we calculated the correlation coefficient between WERs (CER for Korean) on Speech-MASSIVE and FLEURS, resulting in a value of 0.96. This shows that Whisper consistently performs across both datasets, despite Speech-MASSIVE being more challenging than FLEURS for ASR.
\section{SLU Baselines and Beyond}

In this section, we establish several SLU baselines, evaluating them with different training conditions and metrics described in \S\ref{metrics}.  
Firstly (\S\ref{NLU_model_intro}), we build a NLU model, serving as an upper bound free from ASR errors. 
Secondly, we build a cascaded SLU system(\S\ref{Cascaded_SLU_model_intro}), in which an ASR component transcribes input audio and the NLU model utilizes ASR output for inference.
Thirdly, to complete the inventory of SLU baselines, we introduce an end-to-end (E2E) model (\S\ref{E2E_SLU_model_intro}). 
We conclude by showcasing the versatility of Speech-MASSIVE beyond SLU, computing additional baselines for tasks such as speech translation and language identification (\S\ref{more_baseline}).

\subsection{NLU/SLU training conditions and metrics}
\label{metrics}

To simulate different training resource scenarios, we report performance in three different settings: \textit{(a) Zero-shot:}
we train the model only with one language data from the train split (11,514 utterances) and evaluate in all other different languages; \textit{(b) Few-shot:} we employ subsets (115 examples) for each of the 12 non-English languages, aligning with our train-115 split.
\footnote{train-115 covers all 18 domains, 60 intents, and 55 slots  (including empty slots).} 
Additionally, we integrate the full zero-shot training split to enrich the multilingual training dataset, totaling 12.8k samples for training;
\textit{(c) Full fine-tune (NLU only):} 11,514 training examples of all 12 languages are pooled (138k samples for training).
We assess intent prediction in a given text or speech with \textit{intent accuracy}\footnote{Due to space limitations, we report only intent accuracy scores.
However, additional SLU metrics (e.g., micro-averaged slot F1, exact match accuracy, slot-type F1, slot-value CER) exhibit a similar trend and are available in the GitHub repository.
We report the average result (and standard deviation) of three runs with different seeds. All experiments were executed on 1 A100 80GB GPU.}.

\subsection{NLU model}
\label{NLU_model_intro}

Our NLU system uses the mT5 encoder-decoder architecture \cite{mt5}, selected for its superior performance as demonstrated in \cite{massive}, where the mT5 \textit{text-to-text} model outperformed both the mT5 encoder-only model and the XLM-R model \cite{xlm-r}.
We use a pre-trained mT5-base model,\footnote{\url{https://huggingface.co/google/mt5-base}} and fine-tune both the encoder and decoder in a sequence-to-sequence manner.
We supply source and target texts as described in \cite{massive} and shown in Figure \ref{fig:text_format}. 
For instance, the French sentence (\textit{Fr}) \textit{``où puis-je aller ce soir''} is annotated in slots (\textit{Fr-Slots}) as  \textit{`Other Other Other timeofday timeofday''} and intent (\textit{Intent}) as \textit{``recommendation\_events''}
in MASSIVE. 
We adapt those annotations to create source and target texts to be used in training: for the source text (\textit{Fr-Src in [NLU]}), we prepend \textit{``Annotate:''} to the French sentence (\textit{Fr}); for the target text (\textit{Fr-Tgt in [NLU]}), we concatenate slots (\textit{Fr-Slots}) and intent (\textit{Intent}). 

Figure \ref{fig:NLU-SLU} displays the intent accuracy results of our NLU system across all languages and modes (zero-shot, few-shot, full fine-tune), along with those of the cascaded SLU models discussed in \S\ref{Cascaded_SLU_model_intro}.
Unsurprisingly, NLU performance increases when moving from zero-shot to full fine-tune regimes. Also, as expected, higher scores are observed for languages (\textit{Nl}, \textit{Fr}, \textit{De}, \textit{Pt}, \textit{Ru}, \textit{Es} and \textit{Pl}) that are better represented in the mC4 multilingual dataset used to train  mT5 model \cite{mt5}.
Finally, the highest results align with those reported in the MASSIVE paper \cite{massive}, serving as a suitable reference upper bound for comparisons with the SLU models discussed in the following section.

\definecolor{lightgray}{RGB}{220, 220, 220}

\begin{figure}
  \begin{tcolorbox}[
    colback=lightgray,
    colframe=black, 
    arc=2pt, 
    boxrule=1pt, 
    left=-60pt, 
    right=0pt, 
    top=0pt, bottom=0pt
  ]
    \begin{floatrow}
      \ffigbox{\begin{minipage}
      {3in}
        \raggedright
        \fontsize{7}{8}\selectfont
        \textbf{\texttt{[Original text in MASSIVE]}} \\
        \texttt{\textit{En)} where can i go tonight} \\
        \texttt{\textit{En-Annot)} where can i go [timeofday : tonight]} \\
        \texttt{\textit{En-Slots)} Other Other Other Other timeofday} \\
        \vspace{0.2\baselineskip}
        \texttt{\textit{Fr)} où puis-je aller ce soir} \\
        \texttt{\textit{Fr-Annot)} où puis-je aller [timeofday:ce soir]} \\
        \texttt{\textit{Fr-Slots)} Other Other Other timeofday timeofday} \\
        \vspace{0.2\baselineskip}
        \texttt{\textit{Intent)} recommendation\_events} \\
        
        \vspace{0.3\baselineskip}
        
        \textbf{\texttt{[NLU]}} \\
        \texttt{\textit{Fr-Src)} Annotate: où puis-je aller ce soir} \\
        \texttt{\textit{Fr-Tgt)} Other Other Other timeofday timeofday recommendation\_events} \\ 

        \vspace{0.3\baselineskip}
        
        \textbf{\texttt{[Cascaded SLU]}} \\
        \texttt{\textit{Fr-ASR)} où puis je aller ce soir} \\
        \texttt{\textit{Fr-Src)} Annotate: où puis je aller ce soir} \\
        \texttt{\textit{Fr-Tgt)} Other Other Other timeofday timeofday recommendation\_events} \\

        \vspace{0.3\baselineskip}

        \textbf{\texttt{[E2E SLU]}} \\
        \texttt{\textit{Fr-Tgt)} où puis-je aller ce soir | Other Other Other timeofday timeofday | recommendation\_events} \\
      \end{minipage}}{}
    \end{floatrow}%
  \end{tcolorbox}
  \vspace{-0.3cm}
  \caption{Input/Output formatting across NLU/SLU tasks. En: original English text. Fr: French translation in MASSIVE. Annot, Slots and Intent: slot and intent annotation of MASSIVE.}
  \label{fig:text_format}
\end{figure}

\subsection{Cascaded SLU model}
\label{Cascaded_SLU_model_intro}

We develop a cascaded SLU system in which an ASR model based on Whisper-large-v3 transcribes the speech, and the same NLU models of \S\ref{NLU_model_intro} (zero-shot, few-shot, full fine-tune) predict slots and intent from the transcribed texts.

The SLU intent accuracy scores in Figure \ref{fig:NLU-SLU} reveal that processing automatically transcribed utterances introduces performance drops of varying magnitude across the different languages and training modes. 
This is especially notable for languages with lower ASR quality (i.e., higher WER), such as \textit{Ar}, \textit{Hu}, \textit{Ko}, \textit{Tr}, and \textit{Vn}. 
This supports our expectations about the difficulty for the downstream textual NLU component of the SLU cascade to handle unrecoverable transcription errors. 
As a matter of fact, in zero-shot mode, the distance with the text-only upper-bound NLU system is considerably smaller for languages featuring higher ASR quality.
Similar to what we observed for NLU (\S\ref{NLU_model_intro}), 
cascaded SLU performance in few-shot mode improves thanks to the additional multilingual data.
The gains are particularly significant for languages with lesser representation in mT5 model, such as \textit{Tr}, \textit{Vn}, \textit{Ko}, and \textit{Ar}.
Lastly in full fine-tune mode, leveraging a larger multilingual training dataset leads to substantial performance enhancements. 
While the gains are variable, we observe that: \textit{i)} for some languages (i.e. \textit{De}, \textit{Ru}, and \textit{Es}), the gap with the highest results of the textual NLU upper bound shrinks to less than two points, while \textit{ii)} for all languages, the scores are significantly higher than those achieved by the textual NLU models dealing with clean input not only in zero-shot, but also in few-shot mode.

\definecolor{white}{RGB}{255, 255, 255}

\begin{figure}
  \begin{tcolorbox}[
    colback=white,
    colframe=black, 
    arc=2pt, 
    boxrule=1pt, 
    left=-60pt, 
    right=0pt, 
    top=0pt, bottom=0pt
  ]
    \begin{floatrow}
      \ffigbox{\begin{minipage}
      {3in}
        \raggedright
        \fontsize{7}{8}\selectfont
        \textbf{\texttt{ASR}} \\
        \texttt{[<|startofstranscript|>, <|language\_id|>, <|transcribe|>, <|notimestamps|>]} \\

        \vspace{0.2\baselineskip}
        
        \textbf{\texttt{E2E SLU}} \\
        \texttt{[<|startofstranscript|>, <|language\_id|>, <|transcribe|>, <|startoflm|>, <|notimestamps|>]} \\

        \vspace{0.2\baselineskip}
        
        \textbf{\texttt{LID}} \\
        \texttt{[<|startofstranscript|>]} \\

        \vspace{0.2\baselineskip}

        \textbf{\texttt{ST}} \\
        \texttt{[<|startofstranscript|>, <|language\_id|>, <|translate|>, <|notimestamps|>]} \\
      \end{minipage}}{}
    \end{floatrow}%
  \end{tcolorbox}
   \vspace{-0.3cm}
  \caption{Various task control tokens fed to Whisper's decoder.}
  \label{fig:whisper_config}
\end{figure}
\begin{table*}[!htp]\centering
\resizebox{\linewidth}{!}{
\caption{Intent accuracy of cascaded and E2E SLU. Both E2E SLU zero-shot and few-shot models are trained either with initial English train set of \cite{slurp} (En) or with French train set of Speech-MASSIVE (Fr). We exclude French (*) from the average as fr-FR scores are no longer zero/few-shot when French is used as the training language.}\label{tab:cascaded_vs_e2e_all}
\begin{tabular}{c|c|c|c||c|c|c}\toprule

lang 

&\makecell{Casc. (En) \\\textbf{zero-shot}} 
&\makecell{E2E (En)\\\textbf{zero-shot}}
&\makecell{E2E (Fr)\\\textbf{zero-shot}}
&\makecell{Casc. (En) \\\textbf{few-shot}} 
&\makecell{E2E (En)\\\textbf{few-shot}}
&\makecell{E2E (Fr)\\\textbf{few-shot}}
\\\hline
ar 
&49.27 ± 0.90 
&33.04 ± 4.74 
&40.00 ± 2.44 
&54.56 ± 0.73 
&57.71 ± 1.46 
&61.22 ± 1.74\\ 
de 
&76.29 ± 0.14 
&70.68 ± 1.37 
&73.91 ± 0.73 
&78.08 ± 0.50 
&78.64 ± 0.65 
&78.45 ± 0.64\\ 
es 
&75.70 ± 0.19 
&73.12 ± 0.75 
&78.62 ± 0.41 
&78.05 ± 0.33 
&79.79 ± 0.66 
&80.59 ± 0.31\\ 
fr 
&75.61 ± 0.48 
&68.43 ± 2.30 
&\textit{85.87 ± 0.26*} 
&77.56 ± 0.13 
&77.11 ± 0.77 
&\textit{85.93 ± 0.35*}\\ 
hu
&63.43 ± 0.92 
&36.62 ± 1.49 
&42.28 ± 2.20 
&68.70 ± 0.80 
&60.75 ± 2.40 
&63.93 ± 0.19\\ 
ko 
&60.93 ± 0.84 
&57.96 ± 2.26 
&66.09 ± 1.86 
&68.11 ± 0.04 
&72.82 ± 0.23 
&74.09 ± 0.73\\ 
nl 
&78.82 ± 0.45 
&65.17 ± 0.57 
&67.24 ± 1.44 
&78.93 ± 0.34 
&77.49 ± 0.77 
&77.37 ± 0.47\\ 
pl
&74.57 ± 0.37 
&64.82 ± 1.51 
&64.38 ± 1.29 
&76.11 ± 0.39 
&74.85 ± 0.58 
&76.88 ± 1.37\\ 
pt
&73.12 ± 0.49 
&62.91 ± 1.97 
&72.60 ± 1.01 
&77.21 ± 0.65 
&78.15 ± 1.16 
&80.02 ± 0.29\\ 
ru 
&75.96 ± 0.19 
&69.06 ± 1.71 
&74.75 ± 0.28 
&76.96 ± 0.08 
&79.22 ± 0.67 
&79.51 ± 0.26\\ 
tr 
&65.32 ± 0.61 
&47.60 ± 3.08 
&55.08 ± 1.09 
&70.32 ± 0.48 
&69.44 ± 1.62 
&71.14 ± 1.15\\ 
vi
&60.19 ± 0.39 
&35.44 ± 1.48 
&49.67 ± 2.30 
&64.77 ± 0.98 
&63.36 ± 1.69 
&68.71 ± 0.33\\ 
\hline
\textbf{avg.} &\textbf{69.10 ± 0.19} &\textbf{57.07 ± 1.82} &\textbf{62.24 ± 0.92} &\textbf{72.45 ± 0.32} &\textbf{72.45 ± 0.53} &\textbf{73.81 ± 0.58}\\
\bottomrule
\end{tabular}
}
\end{table*}

\subsection{E2E SLU model}
\label{E2E_SLU_model_intro}

To complete the inventory of SLU baselines for comparison, we introduce an end-to-end (E2E) SLU model: a direct solution that bypasses intermediate text representations (ASR transcripts).
We utilize Whisper, following the approach proposed in \cite{whislu}, which showed superior performance compared to cascaded systems and other speech encoders like wav2vec2.0 \cite{wav2vec2} and HuBERT \cite{hubert}.
Model training follows a sequence-to-sequence approach, with predictions extended to include transcript, slots, and intent.
This allows us to leverage both speech and text information in the model's predictions.
We introduce an additional separator \texttt{``|''} between the tasks, allowing Whisper's tokenizer to tokenize the target text as is, without the need to add slots or intents to the original vocabulary.
Two specific tokens, \texttt{``|''} and \texttt{``\_''}, are removed from Whisper's suppressed token list, as they are required for predicting SLU outputs as task separators and in certain intent values.
In zero-shot mode, we fine-tune Whisper-large-v3 with either a) the English train set of \cite{slurp}, or b) the French train set of Speech-MASSIVE.
These  two conditions (\textit{En} \textit{vs} \textit{Fr}) allow us to investigate the impact of the training language on zero-shot E2E SLU across all other languages. 
Additionally, in few-shot mode, we fine-tune Whisper-large-v3 with the English or French train sets, along with train-115 splits from other languages.
We do not provide  a full fine-tune E2E SLU mode since only two languages in Speech-MASSIVE are supported by full train splits.

Table \ref{tab:cascaded_vs_e2e_all} compares cascaded and E2E SLU performance in both zero-shot and few-shot modes.
It is worth noting that the comparison between the two approaches is fair only when using the English train set (\textit{En}), since they utilize the same training utterances albeit in different modalities (written form for cascade and spoken form for E2E).
In this condition (\textit{En}), for zero-shot mode, cascaded SLU outperforms E2E SLU  for all languages. 
In few-shot mode, we note a different trend, with cascaded and E2E models exhibiting similar average performance. 
Employing the French training set from Speech-MASSIVE (\textit{Fr}), E2E SLU surpasses models trained on the English dataset from \cite{slurp} (\textit{En}) in both zero-shot and few-shot modes. 
In zero-shot mode, we observe improvements of more than 5 points for 9 out of 11 languages.
In few-shot mode, although the influence of the training language (\textit{En} \textit{vs} \textit{Fr}) diminishes due to multilingual training, using French as the majority language still yields better performance than using English.
These results highlight the significant influence of the `training language' on the performance of E2E SLU models in zero/few-shot settings. 
Speech-MASSIVE provides a unique opportunity to explore this intriguing observation further.
Finally, examining French (\textit{Fr}) results representing the full fine-tune mode for this language, E2E SLU achieves intent accuracy of 85.87\%, compared to 84.73\% for cascaded SLU and 87.43\% for NLU given in Fig.\ref{fig:NLU-SLU}.

\subsection{Other baselines}

\begin{table*}[!htp]
\centering
\resizebox{\linewidth}{!}{
\caption{LID accuracy and ST BLEU results with Whisper-large-v3 on Speech-MASSIVE.}
\label{tab:lid_st}
\begin{tabular}{c|cc|cc|cc|cc|cc|cc|cc|cc|cc|cc|cc|cc}\toprule
lang &\multicolumn{2}{c|}{ar} &\multicolumn{2}{c|}{de} &\multicolumn{2}{c|}{es} &\multicolumn{2}{c|}{fr} &\multicolumn{2}{c|}{hu} &\multicolumn{2}{c|}{ko} &\multicolumn{2}{c|}{nl} &\multicolumn{2}{c|}{pl} &\multicolumn{2}{c|}{pt} &\multicolumn{2}{c|}{ru} &\multicolumn{2}{c|}{tr} &\multicolumn{2}{c}{vi} \\\hline
split &dev &test &dev &test &dev &test &dev &test &dev &test &dev &test &dev &test &dev &test &dev &test &dev &test &dev &test &dev &test \\
\hline
LID accuracy &90.9 &89.5 &98.9 &98.4 &99.0 &98.6 &98.7 &98.9 &94.6 &95.8 &99.1 &98.7 &94.8 &94.9 &95.3 &94.6 &95.9 &96.0 &99.1 &98.8 &96.1 &96.0 &90.7 &93.2 \\
ST BLEU &17.2 &16.6 &36.7 &38.2 &38.5 &38.2 &38.7 &40.1 &19.4 &20.6 &19.7 &19.5 &40.0 &38.9 &29.9 &28.8 &32.4 &32.3 &28.4 &28.2 &26.7 &26.0 &18.9 &20.2 \\
\bottomrule
\end{tabular}
}
\end{table*}

\label{more_baseline}
We conclude our experiments using  Whisper-large-v3 without any finetuning to compute other baselines and demonstrate the versatility of Speech-MASSIVE. 
We perform Language Identification (LID) and Speech Translation (ST) across \textit{x}$\rightarrow$en language directions.
Different types of tokens are fed to Whisper's decoder depending on the tasks as shown in Figure \ref{fig:whisper_config}.
Table \ref{tab:lid_st} reports Whisper-large-v3 model's LID accuracy and ST BLEU \cite{bleu_metric} on Speech-MASSIVE. 
LID is calculated over all the samples in dev and test splits. 
For ST, instead, BLEU is computed on subsets of dev and test splits identified using meta information from MASSIVE to exclude samples with \textit{localized} translation.
This filtering is necessary to ensure an accurate assessment of translation quality, as localized references may introduce discrepancies in word choice (see \S\ref{introduction}).
Besides indicating the versatility of Speech-MASSIVE for evaluation purposes, our additional baselines on speech-related tasks offer valuable reference scores for cross-task comparisons and for exploring collaborative solutions to leverage potential mutual benefits.

\section{Conclusion}

We introduced Speech-MASSIVE, a multilingual SLU dataset spanning 12 languages for intent prediction and slot-filling tasks. Alongside dataset creation, we established baselines for SLU across various resource and architecture configurations. Additionally, we showcased Speech-MASSIVE's versatility beyond SLU, extending to tasks such as ASR, LID, and ST. With its diverse array of native speakers and recording environments, Speech-MASSIVE holds promise as a benchmark for multilingual, multimodal, and multi-task speech research. Future research opportunities include exploring further the influence of training languages on zero/few-shot SLU performance, thoroughly comparing cascade and E2E SLU solutions, assess the effect of including multi-task and multilingual corpora in the training of speech foundation models, and pushing the boundaries of  E2E multi-task speech systems beyond our baselines.

\section*{Acknowledgements}
The speech collection was funded  by EU Horizon Europe~(HE) Research and Innovation programme grant No 101070631. We also acknowledge the support of the PNRR project FAIR -  Future AI Research (PE00000013),  under the NRRP MUR program funded by the NextGenerationEU.

{
    \small
    \bibliographystyle{ieeenat_fullname}
    \bibliography{main}
}

\clearpage
\appendix
\section{Appendix}
The hyper parameter used to train the end-to-end Spoken Language Understanding model is presented in Table \ref{tab:hparam}.

We report all the evaluation results Exact match accuracy in Table \ref{tab:apdx_ema}, Intent accuracy in Table \ref{tab:apdx_ia}, Slot micro-F1 in Table \ref{tab:apdx_slot_micro_f1} and both Slot type F1 and Slot value CER in Table \ref{tab:apdx_slot_f1_n_cer}.

Evaluation code used to calculate the metrics of Exact match accuracy, Intent accuracy and Slot micro-F1 is from MASSIVE\cite{massive} implementation. \footnote{\url{https://github.com/alexa/massive/blob/main/src/massive/utils/training_utils.py}} For Slot type F1 and slot value CER evaluation, we use S3PRL toolkit.\footnote{\url{https://github.com/s3prl/s3prl/blob/aa3ba844bfe2b5402b7f345cbebd72b33ef6aeff/s3prl/metric/slot_filling.py}}

\begin{table*}[]
\begin{tabular}{c|c}
adam beta1                     & 0.8                                                                          \\ \hline
adam beta2                     & 0.999                                                                        \\ \hline
adam epsilon                   & 0                                                                            \\ \hline
add separator                  & `|'                                                                          \\ \hline
freeze feature encoder         & FALSE                                                                        \\ \hline
gradient accumulation steps    & 2                                                                            \\ \hline
gradient checkpointing         & FALSE                                                                        \\ \hline
label smoothing factor         & 0                                                                            \\ \hline
learning rate                  & 0.00001                                                                      \\ \hline
lr scheduler type              & linear                                                                       \\ \hline
max steps                      & \begin{tabular}[c]{@{}c@{}}20000 (zero-shot)\\ 25000 (few-shot)\end{tabular} \\ \hline
per device eval batch size     & 8                                                                            \\ \hline
per device train batch size    & 8                                                                            \\ \hline
target format content          & transcript slots intent                                                      \\ \hline
task                           & transcribe                                                                   \\ \hline
tokens to remove from suppress & `Ġ|', `\_'                                                                     \\ \hline
warmup steps                   & \begin{tabular}[c]{@{}c@{}}2000 (zero-shot)\\ 2500 (few-shot)\end{tabular}  
\end{tabular}
\caption{Hyper parameter settings for end-to-end SLU zero-shot and few-shot models.}
\label{tab:hparam}
\end{table*}
\begin{table*}[!hb]
\centering
\resizebox{\linewidth}{!}{
\begin{tabular}{c|cccc|cccc|cc}
\multicolumn{1}{l|}{} & \multicolumn{4}{c|}{\textbf{zero-shot}}                                                      & \multicolumn{4}{c|}{\textbf{few-shot}}                                                    & \multicolumn{2}{c}{\textbf{fine-tune}}        \\ \cline{2-11} 
lang                  & NLU                   & Cascaded SLU          & E2E SLU (En)         & E2E SLU (Fr)          & NLU                   & Cascaded SLU          & E2E SLU (En)       & E2E SLU (Fr)         & NLU                   & Cascaded SLU          \\ \hline
ar-SA                 & 28.39 ± 1.16          & 20.93 ± 1.04          & 17.81 ± 2.47         & 22.22 ± 1.72          & 39.48 ± 1.32          & 27.55 ± 0.78          & 37.47 ± 0.30        & 39.76 ± 1.32         & 64.64 ± 0.58          & 45.06 ± 0.29          \\
de-DE                 & 51.18 ± 1.02          & 44.71 ± 1.01          & 46.71 ± 1.43         & 49.61 ± 0.87          & 56.06 ± 1.13          & 48.28 ± 0.54          & 56.44 ± 0.62       & 57.64 ± 0.19         & 69.70 ± 0.53           & 60.35 ± 0.34          \\
es-ES                 & 47.71 ± 0.44          & 44.40 ± 0.60            & 50.00 ± 0.83            & 54.55 ± 0.47          & 52.64 ± 0.39          & 49.23 ± 0.42          & 57.42 ± 0.45       & 59.09 ± 0.36         & 67.09 ± 0.12          & 61.84 ± 0.09          \\
fr-FR                 & 45.09 ± 1.11          & 38.3 ± 0.79           & 43.67 ± 2.09         & 65.38 ± 0.43          & 52.59 ± 1.34          & 38.88 ± 0.14          & 53.04 ± 0.55       & 65.61 ± 0.25         & 67.44 ± 0.17          & 46.08 ± 0.29          \\
hu-HU                 & 38.11 ± 1.25          & 31.46 ± 1.01          & 19.1 ± 0.83          & 22.10 ± 1.46           & 47.00 ± 0.74             & 37.40 ± 0.48           & 38.70 ± 2.60         & 42.57 ± 0.50          & 68.35 ± 0.46          & 53.12 ± 0.12          \\
ko-KR                 & 31.97 ± 0.81          & 30.08 ± 0.81          & 33.55 ± 2.06         & 39.70 ± 1.41           & 43.97 ± 0.20           & 37.73 ± 0.12          & 47.81 ± 0.76       & 49.18 ± 0.95         & 69.45 ± 0.45          & 55.34 ± 0.15          \\
nl-NL                 & 52.05 ± 1.10           & 46.21 ± 1.24          & 40.51 ± 0.34         & 42.17 ± 0.79          & 56.2 ± 0.19           & 47.89 ± 0.05          & 53.78 ± 1.16       & 54.95 ± 0.27         & 69.72 ± 0.21          & 59.18 ± 0.17          \\
pl-PL                 & 45.30 ± 0.61           & 41.32 ± 0.64          & 39.59 ± 1.68         & 40.22 ± 0.91          & 49.22 ± 0.69          & 44.27 ± 0.63          & 51.57 ± 0.69       & 54.75 ± 0.74         & 65.98 ± 0.40           & 58.61 ± 0.44          \\
pt-PT                 & 46.35 ± 0.58          & 39.84 ± 0.74          & 38.49 ± 1.00            & 46.55 ± 1.04          & 52.72 ± 0.87          & 44.48 ± 0.59          & 53.64 ± 0.52       & 56.01 ± 0.26         & 68.84 ± 0.05          & 56.92 ± 0.05          \\
ru-RU                 & 48.9 ± 1.06           & 46.32 ± 1.35          & 45 ± 1.68            & 49.86 ± 0.24          & 53.25 ± 0.34          & 49.78 ± 0.40           & 57.79 ± 0.57       & 58.09 ± 0.07         & 70.17 ± 0.14          & 64.52 ± 0.15          \\
tr-TR                 & 37.88 ± 0.25          & 32.33 ± 0.2           & 24.56 ± 2.25         & 30.14 ± 0.74          & 45.79 ± 0.88          & 37.56 ± 0.50           & 44.98 ± 1.71       & 46.62 ± 0.88         & 68.91 ± 0.32          & 55.51 ± 0.47          \\
vi-VN                 & 30.35 ± 0.93          & 25.96 ± 0.69          & 13.85 ± 0.39         & 21.15 ± 1.77          & 38.25 ± 1.47          & 31.01 ± 1.26          & 35.35 ± 1.12       & 41.35 ± 0.31         & 65.78 ± 0.24          & 51.46 ± 0.45          \\ \hline
\textbf{avg}          & \textbf{41.94 ± 0.67} & \textbf{36.82 ± 0.65} & \textbf{34.4 ± 1.28} & \textbf{38.02 ± 0.84} & \textbf{48.93 ± 0.71} & \textbf{41.17 ± 0.41} & \textbf{49 ± 0.47} & \textbf{50.91 ± 0.4} & \textbf{68.01 ± 0.18} & \textbf{55.66 ± 0.13}
\end{tabular}
}
\caption{Exact Match Accuracy for all the settings.}
\label{tab:apdx_ema}
\end{table*}

\clearpage
\newpage

\begin{table*}[!htp]\centering
\resizebox{\linewidth}{!}{

\begin{tabular}{c|cccc|cccc|cc}
\multicolumn{1}{l|}{} & \multicolumn{4}{c|}{\textbf{zero-shot}}                                                      & \multicolumn{4}{c|}{\textbf{few-shot}}                                                        & \multicolumn{2}{c}{\textbf{fine-tune}}        \\ \cline{2-11} 
lang                  & NLU                   & Cascaded SLU         & E2E SLU (En)          & E2E SLU (Fr)          & NLU                   & Cascaded SLU          & E2E SLU (En)          & E2E SLU (Fr)          & NLU                   & Cascaded SLU          \\ \hline
ar-SA                 & 58.65 ± 0.40           & 49.27 ± 0.90          & 33.04 ± 4.74          & 40.00 ± 2.44             & 65.23 ± 1.23          & 54.56 ± 0.73          & 57.71 ± 1.46          & 61.22 ± 1.74          & 82.47 ± 0.20           & 71.13 ± 0.28          \\
de-DE                 & 77.82 ± 0.14          & 76.29 ± 0.14         & 70.68 ± 1.37          & 73.91 ± 0.73          & 79.49 ± 0.64          & 78.08 ± 0.50           & 78.64 ± 0.65          & 78.45 ± 0.64          & 86.77 ± 0.21          & 85.46 ± 0.19          \\
es-ES                 & 76.61 ± 0.27          & 75.70 ± 0.19          & 73.12 ± 0.75          & 78.62 ± 0.41          & 78.87 ± 0.07          & 78.05 ± 0.33          & 79.79 ± 0.66          & 80.59 ± 0.31          & 87.15 ± 0.38          & 85.87 ± 0.27          \\
fr-FR                 & 78.55 ± 0.44          & 75.61 ± 0.48         & 68.43 ± 2.30           & 85.87 ± 0.26          & 79.91 ± 0.51          & 77.56 ± 0.13          & 77.11 ± 0.77          & 85.93 ± 0.35          & 87.43 ± 0.41          & 84.73 ± 0.35          \\
hu-HU                 & 68.14 ± 0.95          & 63.43 ± 0.92         & 36.62 ± 1.49          & 42.28 ± 2.20           & 72.99 ± 0.61          & 68.70 ± 0.80            & 60.75 ± 2.40           & 63.93 ± 0.19          & 86.42 ± 0.27          & 81.82 ± 0.21          \\
ko-KR                 & 62.34 ± 0.93          & 60.93 ± 0.84         & 57.96 ± 2.26          & 66.09 ± 1.86          & 70.50 ± 0.10            & 68.11 ± 0.04          & 72.82 ± 0.23          & 74.09 ± 0.73          & 86.59 ± 0.29          & 83.40 ± 0.30            \\
nl-NL                 & 80.06 ± 0.42          & 78.82 ± 0.45         & 65.17 ± 0.57          & 67.24 ± 1.44          & 80.4 ± 0.22           & 78.93 ± 0.34          & 77.49 ± 0.77          & 77.37 ± 0.47          & 87.81 ± 0.30           & 85.82 ± 0.40           \\
pl-PL                 & 76.63 ± 0.42          & 74.57 ± 0.37         & 64.82 ± 1.51          & 64.38 ± 1.29          & 77.93 ± 0.28          & 76.11 ± 0.39          & 74.85 ± 0.58          & 76.88 ± 1.37          & 87.53 ± 0.04          & 85.17 ± 0.16          \\
pt-PT                 & 75.81 ± 0.05          & 73.12 ± 0.49         & 62.91 ± 1.97          & 72.60 ± 1.01           & 79.79 ± 0.76          & 77.21 ± 0.65          & 78.15 ± 1.16          & 80.02 ± 0.29          & 87.70 ± 0.26           & 85.24 ± 0.24          \\
ru-RU                 & 77.15 ± 0.33          & 75.96 ± 0.19         & 69.06 ± 1.71          & 74.75 ± 0.28          & 77.92 ± 0.30           & 76.96 ± 0.08          & 79.22 ± 0.67          & 79.51 ± 0.26          & 87.80 ± 0.22           & 86.57 ± 0.27          \\
tr-TR                 & 68.65 ± 0.30           & 65.32 ± 0.61         & 47.60 ± 3.08           & 55.08 ± 1.09          & 73.99 ± 0.37          & 70.32 ± 0.48          & 69.44 ± 1.62          & 71.14 ± 1.15          & 86.97 ± 0.26          & 83.13 ± 0.24          \\
vi-VN                 & 64.55 ± 0.80           & 60.19 ± 0.39         & 35.44 ± 1.48          & 49.67 ± 2.30           & 71.21 ± 1.06          & 64.77 ± 0.98          & 63.36 ± 1.69          & 68.71 ± 0.33          & 86.13 ± 0.16          & 78.17 ± 0.59          \\ \hline
\textbf{avg}          & \textbf{72.08 ± 0.21} & \textbf{69.10 ± 0.19} & \textbf{57.07 ± 1.82} & \textbf{62.24 ± 0.92} & \textbf{75.69 ± 0.42} & \textbf{72.45 ± 0.32} & \textbf{72.45 ± 0.53} & \textbf{73.81 ± 0.58} & \textbf{86.73 ± 0.13} & \textbf{83.04 ± 0.15}
\end{tabular}
}
\caption{Intent Accuracy for all the settings.}
\label{tab:apdx_ia}
\end{table*}

\begin{table*}[!htp]\centering
\resizebox{\linewidth}{!}{

\begin{tabular}{c|cccc|cccc|cc}
\multicolumn{1}{l|}{} & \multicolumn{4}{c|}{\textbf{zero-shot}}                                                       & \multicolumn{4}{c|}{\textbf{few-shot}}                                                        & \multicolumn{2}{c}{\textbf{fine-tune}}        \\ \cline{2-11} 
lang                  & NLU                   & Cascaded SLU          & E2E SLU (En)          & E2E SLU (Fr)          & NLU                   & Cascaded SLU          & E2E SLU (En)          & E2E SLU (Fr)          & NLU                   & Cascaded SLU          \\ \hline
ar-SA                 & 36.49 ± 1.89          & 26.89 ± 1.77          & 10.23 ± 2.64          & 16.41 ± 1.33          & 54.21 ± 1.14          & 40.20 ± 1.07          & 36.27 ± 0.69          & 38.58 ± 0.67          & 76.50 ± 0.52          & 54.66 ± 0.27          \\
de-DE                 & 62.49 ± 0.56          & 55.04 ± 0.21          & 50.56 ± 1.95          & 54.12 ± 1.09          & 69.32 ± 0.76          & 58.92 ± 0.37          & 59.56 ± 0.85          & 62.12 ± 0.54          & 80.00 ± 0.54          & 67.59 ± 0.22          \\
es-ES                 & 56.35 ± 1.06          & 52.09 ± 1.21          & 47.51 ± 0.55          & 50.38 ± 0.80          & 63.35 ± 0.86          & 58.79 ± 0.82          & 55.04 ± 2.03          & 58.39 ± 0.31          & 76.00 ± 0.10          & 69.12 ± 0.20          \\
fr-FR                 & 49.57 ± 0.82          & 37.91 ± 0.45          & 37.17 ± 2.2           & 62.30 ± 0.47          & 62.30 ± 1.81          & 38.71 ± 0.68          & 49.93 ± 0.29          & 62.31 ± 0.22          & 76.38 ± 0.33          & 43.19 ± 0.53          \\
hu-HU                 & 44.47 ± 1.03          & 36.25 ± 1.09          & 18.54 ± 1.51          & 22.61 ± 0.67          & 60.68 ± 0.83          & 47.16 ± 0.68          & 42.01 ± 0.24          & 46.12 ± 0.17          & 78.70 ± 0.19          & 59.01 ± 0.07          \\
ko-KR                 & 42.83 ± 1.52          & 39.29 ± 1.01          & 23.38 ± 3.55          & 27.37 ± 1.78          & 57.20 ± 0.64          & 47.12 ± 0.09          & 41.43 ± 0.89          & 43.30 ± 1.11          & 80.06 ± 0.60          & 60.97 ± 0.29          \\
nl-NL                 & 60.92 ± 0.87          & 52.50 ± 1.39          & 44.57 ± 0.58          & 44.88 ± 0.45          & 68.68 ± 0.34          & 56.29 ± 0.09          & 55.90 ± 0.14          & 57.68 ± 0.30          & 78.33 ± 0.29          & 63.33 ± 0.30          \\
pl-PL                 & 52.69 ± 0.96          & 47.53 ± 1.04          & 35.93 ± 2.55          & 40.74 ± 1.07          & 61.35 ± 0.61          & 54.69 ± 0.36          & 51.53 ± 1.29          & 55.20 ± 0.24          & 74.14 ± 0.49          & 65.29 ± 0.35          \\
pt-PT                 & 54.45 ± 0.87          & 44.50 ± 0.98          & 32.72 ± 1.31          & 40.12 ± 2.22          & 62.49 ± 0.94          & 50.34 ± 0.65          & 48.48 ± 1.46          & 51.13 ± 0.67          & 77.63 ± 0.19          & 60.63 ± 0.34          \\
ru-RU                 & 59.73 ± 1.15          & 55.98 ± 1.00          & 41.10 ± 3.42          & 48.70 ± 1.05          & 64.91 ± 0.71          & 59.62 ± 0.59          & 57.40 ± 0.73          & 59.45 ± 0.47          & 79.08 ± 0.32          & 71.60 ± 0.28          \\
tr-TR                 & 46.74 ± 0.96          & 39.60 ± 0.70          & 24.44 ± 2.76          & 29.51 ± 1.09          & 58.71 ± 0.86          & 47.94 ± 0.18          & 45.95 ± 0.88          & 48.59 ± 0.99          & 78.58 ± 0.55          & 60.85 ± 0.51          \\
vi-VN                 & 36.97 ± 1.13          & 30.24 ± 0.94          & 12.01 ± 2.04          & 21.98 ± 3.06          & 46.02 ± 2.87          & 37.58 ± 2.45          & 44.40 ± 1.50          & 50.26 ± 0.56          & 75.00 ± 0.52          & 58.26 ± 0.45          \\ \hline
\textbf{avg}          & \textbf{50.31 ± 0.86} & \textbf{43.15 ± 0.81} & \textbf{31.51 ± 1.32} & \textbf{36.07 ± 1.04} & \textbf{60.77 ± 0.93} & \textbf{49.78 ± 0.43} & \textbf{48.99 ± 1.50} & \textbf{51.89 ± 1.50} & \textbf{77.53 ± 0.28} & \textbf{61.21 ± 0.15}
\end{tabular}
}
\caption{Micro-avg slot F1 for all the settings.}
\label{tab:apdx_slot_micro_f1}
\end{table*}
\begin{table*}[!htp]\centering
\resizebox{\linewidth}{!}{

\begin{tabular}{c|cccc|cccc}
\multicolumn{1}{l|}{} & \multicolumn{4}{c|}{\textbf{Slot type F1 score}}                                                                   & \multicolumn{4}{c}{\textbf{Slot value CER}}                                                                        \\ \cline{2-9} 
\multicolumn{1}{l|}{} & \multicolumn{2}{c|}{zero-shot}                                     & \multicolumn{2}{c|}{few-shot}                 & \multicolumn{2}{c|}{zero-shot}                                     & \multicolumn{2}{c}{few-shot}                  \\ \cline{2-9} 
lang                  & E2E SLU (En)          & \multicolumn{1}{c|}{E2E SLU (Fr)}          & E2E SLU (En)          & E2E SLU (Fr)          & E2E SLU (En)          & \multicolumn{1}{c|}{E2E SLU (Fr)}          & E2E SLU (En)          & E2E SLU (Fr)          \\ \hline
ar-SA                 & 70.61 ± 1.14          & \multicolumn{1}{c|}{70.70 ± 1.18}          & 78.70 ± 0.50          & 80.14 ± 0.78          & 59.39 ± 2.90          & \multicolumn{1}{c|}{48.75 ± 2.52}          & 33.71 ± 0.21          & 30.81 ± 0.23          \\
de-DE                 & 86.48 ± 0.38          & \multicolumn{1}{c|}{86.65 ± 0.49}          & 89.45 ± 0.67          & 89.71 ± 0.19          & 23.23 ± 1.10          & \multicolumn{1}{c|}{20.74 ± 0.84}          & 16.72 ± 0.34          & 14.59 ± 0.13          \\
es-ES                 & 87.73 ± 0.34          & \multicolumn{1}{c|}{88.81 ± 0.33}          & 89.70 ± 0.24          & 90.25 ± 0.07          & 20.05 ± 0.47          & \multicolumn{1}{c|}{15.61 ± 0.39}          & 14.88 ± 0.63          & 13.42 ± 0.10          \\
fr-FR                 & 86.73 ± 0.82          & \multicolumn{1}{c|}{92.88 ± 0.13}          & 89.62 ± 0.21          & 92.97 ± 0.04          & 25.07 ± 1.25          & \multicolumn{1}{c|}{10.99 ± 0.21}          & 17.40 ± 0.24          & 10.84 ± 0.03          \\
hu-HU                 & 76.20 ± 0.82          & \multicolumn{1}{c|}{77.05 ± 0.45}          & 82.59 ± 0.67          & 83.75 ± 0.32          & 43.71 ± 1.25          & \multicolumn{1}{c|}{42.65 ± 0.66}          & 29.12 ± 1.01          & 25.56 ± 0.05          \\
ko-KR                 & 77.36 ± 2.24          & \multicolumn{1}{c|}{78.62 ± 3.30}          & 84.80 ± 0.37          & 86.24 ± 0.36          & 58.56 ± 3.44          & \multicolumn{1}{c|}{50.55 ± 2.92}          & 36.70 ± 1.86          & 30.89 ± 0.21          \\
nl-NL                 & 86.41 ± 0.28          & \multicolumn{1}{c|}{86.06 ± 0.27}          & 89.11 ± 0.38          & 89.43 ± 0.20          & 27.78 ± 0.69          & \multicolumn{1}{c|}{24.65 ± 0.71}          & 17.76 ± 0.29          & 15.77 ± 0.06          \\
pl-PL                 & 82.83 ± 0.83          & \multicolumn{1}{c|}{83.48 ± 0.34}          & 87.65 ± 0.15          & 88.30 ± 0.25          & 28.77 ± 0.93          & \multicolumn{1}{c|}{28.70 ± 0.47}          & 20.58 ± 0.58          & 17.90 ± 0.25          \\
pt-PT                 & 83.27 ± 0.57          & \multicolumn{1}{c|}{86.23 ± 0.52}          & 89.18 ± 0.30          & 89.94 ± 0.14          & 31.32 ± 1.11          & \multicolumn{1}{c|}{25.04 ± 1.63}          & 17.62 ± 0.36          & 15.78 ± 0.13          \\
ru-RU                 & 84.49 ± 1.00          & \multicolumn{1}{c|}{85.47 ± 0.58}          & 88.76 ± 0.31          & 89.62 ± 0.20          & 36.53 ± 2.05          & \multicolumn{1}{c|}{25.28 ± 1.01}          & 17.91 ± 0.66          & 15.72 ± 0.06          \\
tr-TR                 & 77.86 ± 1.75          & \multicolumn{1}{c|}{78.55 ± 1.93}          & 84.82 ± 0.89          & 85.78 ± 0.47          & 37.96 ± 1.59          & \multicolumn{1}{c|}{35.56 ± 1.56}          & 24.43 ± 0.41          & 21.57 ± 0.44          \\
vi-VN                 & 75.41 ± 0.30          & \multicolumn{1}{c|}{78.67 ± 1.31}          & 83.60 ± 0.91          & 85.56 ± 0.40          & 52.96 ± 0.92          & \multicolumn{1}{c|}{41.99 ± 1.96}          & 30.10 ± 0.57          & 26.57 ± 0.26          \\ \hline
\textbf{avg}          & \textbf{81.28 ± 0.78} & \multicolumn{1}{c|}{\textbf{81.84 ± 0.90}} & \textbf{86.50 ± 0.18} & \textbf{87.16 ± 1.50} & \textbf{37.11 ± 0.98} & \multicolumn{1}{c|}{\textbf{32.68 ± 1.09}} & \textbf{23.08 ± 0.21} & \textbf{20.78 ± 1.50}
\end{tabular}
}
\caption{Slot type F1 score and slot value CER for all the settings.}
\label{tab:apdx_slot_f1_n_cer}
\end{table*}

\end{document}